\documentclass[11pt,a4paper]{article} 
\usepackage[hyperindex,breaklinks]{hyperref}
\usepackage[hyperref]{emnlp2020}
\usepackage{times}
\usepackage{latexsym}

\usepackage{microtype}

\usepackage{booktabs,adjustbox,amsmath,amssymb}
\usepackage{xcolor,colortbl,multirow}
\usepackage{tikz,tikz-qtree}

\DeclareMathOperator*{\argmax}{arg\,max}

\aclfinalcopy 
\def\aclpaperid{1494} 

\title{Unsupervised Parsing via Constituency Tests}

\author{Steven Cao \qquad Nikita Kitaev \qquad Dan Klein \\
	Computer Science Division \\
	University of California, Berkeley \\
	{\tt \{stevencao,kitaev,klein\}@berkeley.edu}}

\date{}

\begin{document}
	\maketitle
	\begin{abstract}
		We propose a method for unsupervised parsing based on the linguistic notion of a constituency test. One type of constituency test involves modifying the sentence via some transformation (e.g.\ replacing the span with a pronoun) and then judging the result (e.g.\ checking if it is grammatical). Motivated by this idea, we design an unsupervised parser by specifying a set of transformations and using an unsupervised neural acceptability model to make grammaticality decisions. To produce a tree given a sentence, we score each span by aggregating its constituency test judgments, and we choose the binary tree with the highest total score. While this approach already achieves performance in the range of current methods, we further improve accuracy by fine-tuning the grammaticality model through a refinement procedure, where we alternate between improving the estimated trees and improving the grammaticality model. The refined model achieves 62.8 F1 on the Penn Treebank test set, an absolute improvement of 7.6 points over the previous best published result.
	\end{abstract}
	
	\section{Introduction}
	
	When developing a phrase structure grammar, one powerful tool that linguists use is constituency tests. Given a sentence and a span within it, one type of constituency test involves modifying the sentence via some transformation (e.g.\ replacing the span with a pronoun) and then judging the result (e.g.\ checking if it is grammatical). If a span passes constituency tests, then linguists have evidence that it is a constituent. Motivated by this idea, as well as recent advancements in neural acceptability (grammaticality) models via pre-training~\citep{warstadt2018neural,devlin_bert:2018,liu-etal-2019-roberta}, in this paper we propose a method for unsupervised parsing that operationalizes the way linguists use constituency tests. 
	
	Focusing on constituency tests that are judged via grammaticality, we begin by specifying a set of transformations that take as input a span within a sentence and output a new sentence (Section~\ref{section:constituencytests}). Given these transformations, we then describe how to use a (possibly noisy) grammaticality model for parsing (Section~\ref{section:parsing}). Specifically, we score the likelihood that a span is a constituent by applying the constituency tests and averaging their grammaticality judgments, i.e.\ the probability that the transformed sentence is grammatical under the model. We then parse via minimum risk decoding, where we score each binary tree by summing the scores of its contained spans, with the interpretation of maximizing the expected number of constituents. Importantly, this scoring system accounts for false positives and negatives by allowing some spans in the tree to have low probability if the model is confident about the rest of the tree.
	
	To learn the grammaticality model, we note that given gold trees, we can train the model to accept constituency test transformations of gold constituents and reject those of gold distituents. On the other hand, given the model parameters, we can estimate trees via the parsing algorithm in Section~\ref{section:parsing}. Therefore, we learn the model via alternating optimization. First, we learn an initial model by fine-tuning BERT on unlabeled data to distinguish between real sentences and distractors produced by random corruptions like shuffling (Section~\ref{section:grammaticality}). Then, we refine the model by alternating between (1) producing trees, and (2) maximizing/minimizing the scores of predicted constituents/distituents in those trees (Section~\ref{section:em}). 
	
	To evaluate our approach, we compare to existing methods for unsupervised parsing (Section~\ref{section:results}). Our refined model achieves 62.8 F1 averaged over four random restarts on the Penn Treebank (PTB) test set, 7.6 points above the previous best published result, showing that constituency tests provide powerful inductive bias. Analyzing our parser (Section~\ref{section:analysis}), we find that despite its strong numbers, it makes some mistakes that we might expect from the parser's reliance on this class of constituency tests, like attaching modifying phrases incorrectly. As one possible solution to these shortcomings, we use our method to induce the unsupervised recurrent neural network grammar (URNNG)~\citep{kim-etal-2019-unsupervised} following the approach in \citet{kim-etal-2019-compound}, where we use our induced trees as supervision to initialize the RNNG model and then perform unsupervised fine tuning via language modeling. The resulting model achieves 67.9 F1 averaged over four random restarts, approaching the supervised binary tree RNNG with a gap of 4.9 points.
	
	\begin{table*}[t]
		\begin{center}
			\resizebox{\linewidth}{!}{\begin{tabular}{lll}
					\toprule
					Name & Applied to ``A [ B ] C'' & Example \\
					\midrule
					Clefting & it \{is, was\} B that A C & \textit{it \{is, was\} the london market that by midday , was in full retreat} \\
					Coordination & A B and B C & \textit{by midday , the london market and the london market was in full retreat} \\
					Substitution & A \{it, ones, did so\} C & \textit{by midday , \{it, ones, did so\} was in full retreat} \\
					Front Movement & B , A C & \textit{the london market , by midday , was in full retreat} \\
					End Movement & A C B & \textit{by midday , was in full retreat the london market} \\
					\bottomrule
			\end{tabular}}
		\end{center}
		\caption{\label{table:constests} The constituency tests we use in this paper, using the span \textit{``by midday , }[\textit{ the london market }]\textit{ was in full retreat''} as an example.}
	\end{table*}
	
	\begin{table*}[t]
		\begin{center}
			\resizebox{\linewidth}{!}{\begin{tabular}{ll}
					\toprule
					Name & Description \\
					\midrule
					Shuffle & Choose a random subset of words in the sentence and randomly permute them. \\
					Swap & Choose two words and swap them. \\
					Drop & Choose a random subset of words in the sentence and drop them. \\
					Span Drop & Choose a random contiguous span of words and drop it. \\
					Span Movement & Choose a random contiguous span of words and move it to the front or back. \\
					Bigram & Generate a sentence of the same length using a bigram language model trained on the source corpus. \\
					\bottomrule
			\end{tabular}}
		\end{center}
		\caption{\label{table:corruptions} The corruptions we use to train the initial grammaticality model using unlabeled data, where the model must determine whether a given sentence is real or corrupted.}
	\end{table*}
	
	\section{Related Work}\label{section:related}
	
	\textbf{Grammar induction.} There has been a long history of research on grammar induction. Here, we touch on just a couple threads of work most related to our method. Early works focused on building probabilistic context-free grammars (PCFGs) but found that inducing them with expectation-maximization (EM) did not produce meaningful trees~\citep{carroll-charniak-1992-experiments}. We highlight some themes since then that have produced successful unsupervised parsers.
	
	\textit{Directly modeling spans rather than mediating structure through a grammar:} In contrast with previous work based on probabilistic grammars, the constituent-context model of \citet{klein-manning-2002-generative} proposed a probabilistic formulation that modeled the constituency of each span directly, where each span yielded words conditioned on whether or not it was a constituent. Parsing then proceeded via minimum risk decoding~\citep{smith-eisner-2006-minimum}, where they chose the tree with the maximum expected number of constituents.
	
	\textit{Explicitly defining criteria for what it means to be a constituent:} Rather than designing a generative model over sentences and trees, \citet{clark-2001-unsupervised} proposed to identify constituents based on their span statistics, e.g.\ mutual information between left and right contexts of the span.
	
	\textit{Finding external signals of constituency:} To perform noun compound bracketings (``[ \emph{liver cell} ] \emph{line}'' vs ``\emph{liver} [ \emph{cell line} ]''), \citet{nakov-hearst-2005-search} extracted a series of features from Web text, like the frequency of ``\emph{liver-cell line}'' vs ``\emph{liver cell-line}.'' With a similar idea of extracting signal from Web text, \citet{spitkovsky-etal-2010-profiting} found evidence for constituency from HTML markup, e.g.\ hyperlinks and italicized phrases.
	
	\textit{Designing neural latent variable models:} Many works have taken the approach of designing a neural language model with tree-valued latent variables and optimizing it via EM, some of which can also be seen as probabilistic grammars parameterized by neural networks. For example, the compound PCFG~\citep{kim-etal-2019-compound}, found that the original PCFG is sufficient to induce trees if it uses a neural parameterization, and they further enhanced the model via latent sentence vectors to reduce the independence assumptions. Another model, the unsupervised recurrent neural network grammar (URNNG)~\citep{kim-etal-2019-unsupervised}, uses variational inference over latent trees to perform unsupervised optimization of the RNNG~\citep{dyer-etal-2016-recurrent}, an RNN model that defines a joint distribution over sentences and trees via shift and reduce operations. Unlike the PCFG, the URNNG makes no independence assumptions, making it more expressive but also harder to induce from scratch. \citet{shen-etal-2018-neural} proposed the Parsing-Reading-Predict Network (PRPN), where the latent tree structure determines the flow of information in a neural language model, and they found that optimizing for language modeling produced meaningful latent trees. On the other hand, the Deep Inside-Outside Recursive Autoencoder (DIORA)~\citep{drozdov-etal-2019-unsupervised} computes a representation for each node in a tree by recursively combining child representations following the structure of the inside-outside algorithm, and it optimizes an autoencoder objective such that the representation for each leaf in the tree remains unchanged after an inside and outside pass.
	
	\textit{Extracting trees from neural language models:} The Ordered Neuron (ON) model~\citep{shen-etal-2019-ordered} extracts trees from a modified LSTM language model, with the idea that the forget operation typically happens at phrase boundaries. They parse by recursively finding splitpoints based on each neuron's decision of where to forget. More recently, \citet{kim-etal-2020-are} extract trees from pre-trained transformers. Using the model's representations for each word in the sentence, they score fenceposts (positions between words) by computing distance between the two adjacent words, and they parse by recursively splitting the tree at the fencepost with the largest distance.
	
	\noindent\textbf{Neural grammaticality models.} Pre-training has recently produced large gains on a wide range of tasks, including the task of judging whether a sentence is grammatical~\citep{devlin_bert:2018,liu-etal-2019-roberta}. Most works evaluate on the Corpus of Linguistic Acceptability (CoLA)~\citep{warstadt2018neural}, which compiles acceptable and unacceptable sentences from linguistics publications. The paper also investigates the question of whether grammaticality can be learned from unlabeled data, where fake sentences are generated via either random shuffling or an LSTM language model, and the model must determine whether a given sentence is real or fake. They find that real/fake models perform comparably to supervised models trained on the CoLA training set. \citet{lau-etal-2017-grammaticality} also investigate unsupervised acceptability models, where they instead augment language models with a variety of acceptability measures, e.g.\ perplexity renormalized to remove the influence of unigram frequency. They find that such models achieve an encouraging level of agreement with crowd-sourced human judgments.
	
	\section{Constituency Tests}\label{section:constituencytests}
	
	We begin by specifying a set of constituency tests. The constituency tests we focus on involve transformation functions ${c: (\texttt{sent}, i, j) \mapsto \texttt{sent}'}$ that take in a span and output a new sentence, and a judgment function ${g: \texttt{sent} \mapsto \{0, 1\}}$ that judges the resulting sentence. A span passes a constituency test if the judgment function approves of the transformed sentence, or ${g(c(\texttt{sent}, i, j)) = 1}$. Then, parsing via constituency tests involves specifying a set of transformation functions (this section), learning the judgment function (Sections~\ref{section:grammaticality}~and~\ref{section:em}), and aggregating these test results to produce a tree (Section~\ref{section:parsing}). 
	
	We will focus on constituency tests that are judged via grammaticality because it is feasible to learn a grammaticality model using unlabeled data. We describe the set of transformations in Table~\ref{table:constests}. As future work, modeling semantic preservation could also prove fruitful as a way to correct some false positives, e.g.\ \textit{``stock }[\textit{ prices rose after the announcement }]\textit{''} $\to$ \textit{``stock it.''}
	
	Because we specify constituency tests, while the parser is unsupervised in that it doesn't use labeled data, it is not tabula rasa in that we provide it with linguistically-inspired inductive bias, in contrast with past methods that may have less inductive bias or encode it more implicitly. To induce more and specify less, an interesting line of future work would involve inducing the tests as well.
	
	\section{Parsing Algorithm}\label{section:parsing}
	
	With this set of transformations, in this section we describe how to parse sentences using a (potentially noisy) grammaticality model. In the supervised setting, \citet{stern-etal-2017-minimal} and \citet{kitaev-klein-2018-constituency} showed that independently scoring each span and then choosing the tree with the best total score produced a very accurate and simple parser, while \citet{klein-manning-2002-generative} showed a similar result in the unsupervised setting. Therefore, we also use a span-based approach. 
	
	We will use $g_\theta: \texttt{sent} \mapsto [0,1]$ to denote the grammaticality model with parameters $\theta$, which outputs the probability that a given sentence is grammatical. First, we score each span by averaging the grammaticality judgments of its constituency tests, or
	\begin{equation*}
	s_\theta(\texttt{sent}, i, j) = \frac{1}{|C|} \sum_{c \in C} g_\theta(c(\texttt{sent},i,j)),
	\end{equation*}
	where $C$ denotes the set of constituency tests. Then, we score each tree by summing the scores of its spans and choose the highest scoring binary tree via CKY, or
	\begin{equation*}
	t^*(\texttt{sent}) = \argmax_{t \in T(\texttt{len}(\texttt{sent}))} \sum_{(i,j) \in t} s_\theta(\texttt{sent}, i, j),
	\end{equation*}
    where $T(\texttt{len}(\texttt{sent}))$ denotes the set of binary trees with $\texttt{len}(\texttt{sent})$ leaves. If we interpret the score $s_\theta(\texttt{sent}, i, j)$ as estimating the probability that the span $(\texttt{sent}, i, j)$ is a constituent, then this formulation corresponds to choosing the tree with the highest expected number of constituents, i.e.\ minimum risk decoding~\citep{smith-eisner-2006-minimum}. This scoring system accounts for noisy judgments, which lead to false positives and negatives, by allowing some spans to have low probability if the model is confident about the rest of the tree.
	
	If we want $s_\theta(\texttt{sent}, i, j)$ to estimate the posterior probability that the span is a constituent given the judgments of its constituency tests, or
	\begin{align*}
	\mathbf{P}((\texttt{sent}, i, j)\ &\text{is a constituent} \mid \\
	&\{ g_\theta(c(\texttt{sent},i,j)) : c \in C\} ),
	\end{align*}
	then we might want to do something more sophisticated than taking the average. However, we find that the average performs well while being both parameter-less and simple to interpret, so we leave this avenue of exploration to future work.
	
	\section{Initializing the Grammaticality Model}\label{section:grammaticality}
	
	In this section and the next, we describe how we learn the grammaticality model. Given gold trees, we can train the model to accept constituency test transformations of gold constituents and reject those of gold distituents. On the other hand, given model parameters, we can estimate trees using the parsing algorithm in Section~\ref{section:parsing}. Therefore, we first initialize the model (this section), and we then refine it via alternating optimization (Section~\ref{section:em}).
	
	Previously, \citet{warstadt2018neural} found that LSTM grammaticality models trained with supervision versus those trained on a real/fake task achieved similar correlation with human judgments when evaluating on the Corpus of Linguistic Acceptability (CoLA), a dataset with examples of acceptable and unacceptable taken from linguistic publications. Given an unlabeled corpus of sentences and a set of corruptions, the real/fake task involves predicting whether a given sentence is real or corrupted. Motivated by their result, we train our model via a real/fake task but a wider range of corruptions, as described in Table~\ref{table:corruptions}.
	
	Rather than training from scratch, we fine-tune the RoBERTa model~\citep{liu-etal-2019-roberta}, a BERT variant pre-trained on masked word prediction and next sentence prediction. As our unlabeled sentences, we use 5 million sentences from English Gigaword~\citep{graff-cieri-2003-gigaword}, and we do not perform any early stopping. We report optimization hyperparameters in the appendix.
	
	Comparing the real/fake RoBERTa model to a supervised version, we find that the former achieves 0.21 MCC (Matthews Correlation Coefficient) on the CoLA development set, while the latter achieves 0.73 MCC, in contrast with the finding in \citet{warstadt2018neural} that real/fake and supervised LSTMs achieved similar accuracy (both around 0.2 to 0.3 MCC).\footnote{We did not optimize the corruption set for CoLA MCC.} This gap is not totally surprising given how high the supervised RoBERTa numbers are. However, when used for parsing via constituency tests, the real/fake RoBERTa model outperforms the supervised model by about 6 F1 (before refinement), likely because invalid constituency tests look more like random corruptions than examples from the CoLA training set, which are taken from linguistics publications.
	
	\section{Refining the Grammaticality Model}\label{section:em}
	
	While the unrefined grammaticality model achieves 48.2 F1, which is in the range of current methods (Table~\ref{table:ptb}), we further improve accuracy via alternating optimization, which proceeds as follows:
	\begin{enumerate}
		\item Using the span-based algorithm in Section~\ref{section:parsing}, parse a batch $B$ of sentences to produce trees.
		\item Use these trees as pseudo-gold labels to update the span judgments. Specifically, for each sentence, minimize the loss function
		\begin{align*}
		&\phantom{+} \sum_{(i,j) \in t^*(\texttt{sent})} \log(s_\theta(\texttt{sent},i,j))  \\
		&+ \sum_{(i,j) \not\in t^*(\texttt{sent})} \log(1 - s_\theta(\texttt{sent},i,j)),
		\end{align*}
		i.e.\ binary cross-entropy on each span with inclusion into the predicted tree as the label, summed over the sentences in the batch. 
		
		Note that the span scores $s_\theta(\texttt{sent},i,j))$ are derived from grammaticality judgments of constituency tests, so the only parameters are those in the grammaticality model. Therefore, this step can be thought of as increasing the grammaticality judgment of every constituency test applied to every predicted constituent, while decreasing the judgments for predicted distituents.
		\item Repeat for the next batch of sentences.
	\end{enumerate}
	
	This step can be thought of as encouraging self-consistency between the model's grammaticality judgments and the trees that result from them. For example, CKY might choose a tree where a few of the spans are considered invalid if the model is confident about the other spans in the tree. The refinement procedure would then increase the probability of these initially invalid spans, which might help the model catch spans that it initially missed. We see evidence of this effect in Section~\ref{section:analysis}. In addition, there is an inherent mismatch between the real/fake task that the model was trained on and the constituency test judgment task it is being used for. For example, many of the sentences resulting from constituency tests are far out of distribution from sentences seen during training. Therefore, this step can also be thought of as helping the grammar model adapt to its new setting.
	
	One problem, however, is that the loss function takes a gradient through the grammaticality judgments of all of the constituency tests for every span in the sentence. This computation takes up too much memory, given that a length-30 sentence has about 400 spans and thus about 3000 constituency tests. Therefore, to reduce memory usage, for every sentence we only take the gradient through 16 of the constituency tests, chosen randomly. 
	
	While early stopping would likely improve performance, we instead perform refinement for a fixed number of iterations because we don't have access to labeled data. Specifically, we perform refinement for one epoch on 5000 sentences from the PTB training set (sections 2 to 21), combined with the 2416 sentences in the PTB test set (section~23). We find that the training curve is relatively consistent across runs. We use the same optimization parameters as the ones for the real/fake task, as described in the Appendix.
	
	\section{Results}\label{section:results}
	
	\begin{table}[t]
		\begin{center}
			\resizebox{\linewidth}{!}{\begin{tabular}{lccc}
					\toprule
					& \multicolumn{2}{c}{PTB F1} \\
					Model & Mean & Max \\
					\midrule
					PRPN$^\dagger$~\citep{shen-etal-2018-neural} & 37.4 & 38.1 \\
					URNNG~\citep{kim-etal-2019-unsupervised} & -- & 45.4 \\
					ON$^\dagger$~\citep{shen-etal-2019-ordered} & 47.7 & 49.4 \\
					Neural PCFG$^\dagger$~\citep{kim-etal-2019-compound} & 50.8 & 52.6 \\
					DIORA~\citep{drozdov-etal-2019-unsupervised} & --& 58.9 \\
					Compound PCFG$^\dagger$~\citep{kim-etal-2019-compound} & 55.2 & 60.1 \\
					\midrule
					Left Branching & 8.7 \\
					Balanced & 18.5 \\
					Right Branching & 39.5 \\
					\midrule
					Ours (before refinement) & 48.2 \\
					Ours (after refinement) & \textbf{62.8} & \textbf{65.9} \\
					\midrule
					Oracle Binary Trees & 84.3 \\
					\bottomrule
			\end{tabular}}
		\end{center}
		\caption{\label{table:ptb} Unlabeled sentence-level F1 on the PTB test set without punctuation or unary chains. ``Before refinement'' denotes the parser using the acceptability model after real/fake training, which we only run once. Starting from this initial model, we report the mean and maximum score out of 4 random restarts of refinement. Baseline numbers are taken from \citet{kim-etal-2019-compound}. After refinement, the parser outperforms the previous best method by 7.6 points. \\
		$\dagger$ denotes models trained without punctuation.}
	\end{table}
	
	\begin{table}[t]
		\begin{center}
			\resizebox{\linewidth}{!}{\begin{tabular}{lccc}
					\toprule
					 & \multicolumn{2}{c}{PTB F1} \\
					Model & Initial (Max) & +URNNG \\
					\midrule
					PRPN & 47.9 & 51.6 \\
					ON & 50.0 & 55.1 \\
					Neural PCFG & 52.6 & 58.7 \\
					Compound PCFG & 60.1 &  66.9 \\
					\midrule
					Ours (after refinement) & 65.9 & \textbf{71.3} \\
					\midrule
					Supervised Binary RNNG & 71.9 & 72.8 \\
					\bottomrule
			\end{tabular}}
		\end{center}
		\caption{\label{table:urnng} Unlabeled sentence-level F1 on the PTB test set without punctuation or unary chains. Following the experimental setup in \citet{kim-etal-2019-compound}, ``Initial (Max)'' denotes the induced trees resulting from running the method four times and selecting the best result. Next, we use the induced trees as supervision for RNNG and then run unsupervised RNNG fine-tuning, denoted by the ``+URNNG'' column. ``Supervised Binary RNNG'' denotes training the RNNG on binarized gold trees. Baseline numbers are taken from \citet{kim-etal-2019-compound}. When selecting the best parser out of four runs, our method combined with URNNG approaches the supervised binary RNNG, with a gap of 1.5 points. Departing from the setup of \citet{kim-etal-2019-compound}, we also induced URNNG three more times using the other three runs, which resulted in a mean score of 67.9 across the four runs and a minimum of 61.1.}
	\end{table}
	
	\subsection{F1 on the Penn Treebank}
	
	For evaluation, we report the F1 score with respect to gold trees in the Penn Treebank test set (section 23). Following prior work~\citep{kim-etal-2019-compound,shen-etal-2018-neural,shen-etal-2019-ordered}, we strip punctuation and collapse unary chains before evaluation, and we calculate F1 ignoring trivial spans. The averaging is sentence-level rather than span-level, meaning that we compute F1 for each sentence and then average over all sentences. Because most unsupervised parsing methods only consider fully binary trees, we include the oracle binary tree ceiling, produced by taking the (often flat) gold trees and binarizing them arbitrarily.
	
	Table~\ref{table:ptb} displays the F1 numbers for our method compared to existing unsupervised parsers, where we report mean, max, and min out of four random restarts. Before refinement, at 48.2 F1, the parser is already in the range of existing methods. After refinement, the parser achieves 62.8 F1 averaged over four runs, outperforming the previous best result by 7.6 points.\footnote{While other methods do not report the minimum, our minimum score was 60.4 F1. We also evaluate in the setting where the test set sentences are not available during refinement, and we find similar results (mean: 62.8, max: 64.6, min: 61.5).}
	
	\subsection{Inducing URNNG}
	
	\citet{kim-etal-2019-compound} found that while URNNG (described in Section~\ref{section:related}) fails to outperform right-branching trees on average when trained from scratch, it achieves very good performance when initialized using another method's induced trees. Specifically, they first train RNNG using the induced trees from another method as supervision. Then, they perform unsupervised fine-tuning with a language modeling objective. They find that this procedure produces substantial gains when combined with existing unsupervised parsers.
	
    Following their experimental setup, we use our best parser out of four runs to parse both the PTB training set and test set, and we induce URNNG using these predicted trees. We use the default parameters in the \citet{kim-etal-2019-unsupervised} github, which we report in the Appendix. Table~\ref{table:urnng} shows the resulting F1 on the PTB test set. After URNNG, we achieve 71.3 F1, approaching the performance of the supervised binary RNNG + URNNG with a gap of 1.5 points. However, selecting the best parser out of four requires labeled data, so we also induce URNNG from each of the three other parsers. We find that the mean score across the four runs is 67.9. To close the gap between the max and mean across the four runs, ensembling might be an effective approach; we leave this direction to future work.
    
    One possible reason for why URNNG helps is that the URNNG model makes no independence assumptions, making it very expressive but also also difficult to induce from scratch. Therefore, we can think of this method as removing some of the independence assumptions and other biases of the original model once they have sufficiently guided the unsupervised training.
	
	\section{Analysis}\label{section:analysis}
	
	\begin{table}[t] 
		\begin{center}
			\resizebox{\linewidth}{!}{\begin{tabular}{lccc}
					\toprule
					& Before & After & Best parser \\
					& refinement & (best parser) & + URNNG\\
					\midrule
					SBAR & 0.229 & 0.661 & \textbf{0.853} \\
					NP & 0.604  & 0.794 & \textbf{0.843} \\
					VP & 0.325 & 0.682 & \textbf{0.808} \\
					PP & 0.571 & \textbf{0.862} & 0.844 \\
					ADJP & \textbf{0.664} & 0.626 & 0.556 \\
					ADVP & 0.620 & \textbf{0.639} & 0.546\\
					\midrule
					F1 & 48.2 & 65.9 & \textbf{71.3} \\
					\bottomrule
			\end{tabular}}
		\end{center}
		\caption{\label{table:recallbylabel} Recall by label, or the fraction of gold constituents predicted to be constituents by each model, along with F1 (calculated over all spans). We report numbers for the parser before refinement, the best parser out of four runs of refinement, and URNNG induced from the best parser. Refinement and URNNG both produce large improvements for all categories except ADJPs and ADVPs. }
	\end{table}
	
	\begin{table}[t] 
		\begin{center}
			\resizebox{\linewidth}{!}{\begin{tabular}{lccccccccc}
					\toprule
					& \multicolumn{2}{c|}{Clefting} & \multicolumn{3}{c|}{Proform Substitution} & \multicolumn{2}{c|}{Movement} & Coord-\\
					& is & \multicolumn{1}{c|}{was} & ones & did so & \multicolumn{1}{c|}{it} & front & \multicolumn{1}{c|}{end} & ination \\
					\midrule
					\midrule
					\multicolumn{2}{l}{Before Refinement} \\
					\midrule
					SBAR & \cellcolor{orange!25}0.294 & \cellcolor{orange!25}0.260 & \cellcolor{yellow!0}0.113 & \cellcolor{yellow!0}0.130 & \cellcolor{yellow!0}0.146 & \cellcolor{orange!25}0.279 & \cellcolor{yellow!25}0.319 & \cellcolor{green!40}0.942 \\
					NP & \cellcolor{yellow!25}0.353 & \cellcolor{yellow!25}0.347 & \cellcolor{lime!50}0.458 & \cellcolor{orange!25}0.178 & \cellcolor{lime!50}0.555 & \cellcolor{yellow!0}0.055 & \cellcolor{yellow!0}0.048 & \cellcolor{green!40}0.934 \\
					VP & \cellcolor{yellow!0}0.091 & \cellcolor{yellow!0}0.089 & \cellcolor{yellow!0}0.067 & \cellcolor{lime!50}0.479 & \cellcolor{yellow!0}0.127 & \cellcolor{yellow!0}0.060 & \cellcolor{yellow!0}0.144 & \cellcolor{green!40}0.944 \\
					PP & \cellcolor{yellow!25}0.427 & \cellcolor{yellow!25}0.412 & \cellcolor{orange!25}0.238 & \cellcolor{orange!25}0.165 & \cellcolor{orange!25}0.154 & \cellcolor{yellow!25}0.308 & \cellcolor{lime!50}0.606 & \cellcolor{green!40}0.906 \\
					ADJP & \cellcolor{yellow!25}0.383 & \cellcolor{yellow!25}0.361 & \cellcolor{orange!25}0.286 & \cellcolor{orange!25}0.241 & \cellcolor{yellow!25}0.346 & \cellcolor{yellow!0}0.127 & \cellcolor{orange!25}0.172 & \cellcolor{green!40}0.911 \\
					ADVP & \cellcolor{yellow!25}0.396 & \cellcolor{yellow!25}0.395 & \cellcolor{yellow!0}0.143 & \cellcolor{orange!25}0.185 & \cellcolor{orange!25}0.198 & \cellcolor{orange!25}0.290 & \cellcolor{yellow!25}0.307 & \cellcolor{green!40}0.893 \\
					\midrule
					Distituent & \cellcolor{yellow!0}0.066 & \cellcolor{yellow!0}0.063 & \cellcolor{red!30}0.184 & \cellcolor{yellow!0}0.098 & \cellcolor{yellow!0}0.123 & \cellcolor{yellow!0}0.033 & \cellcolor{yellow!0}0.052 & \cellcolor{red!50}0.456 \\
					\midrule
					F1 & 20.8 & 20.9 & 10.9 & 13.2 & 17.8 & 12.3 & 13.6 & 16.1 \\
					\midrule
					\midrule
					\multicolumn{2}{l}{After Refinement} \\
					\midrule
					SBAR & \cellcolor{orange!25}0.237 & \cellcolor{orange!25}0.223 & \cellcolor{orange!25}0.237 & \cellcolor{green!40}0.770 & \cellcolor{yellow!25}0.374 & \cellcolor{orange!25}0.250 & \cellcolor{orange!25}0.225 & \cellcolor{lime!50}0.539 \\
					NP & \cellcolor{green!40}0.718 & \cellcolor{green!40}0.712 & \cellcolor{lime!50}0.571 & \cellcolor{yellow!25}0.428 & \cellcolor{lime!50}0.539 & \cellcolor{yellow!0}0.063 & \cellcolor{yellow!0}0.035 & \cellcolor{green!40}0.792 \\
					VP & \cellcolor{yellow!0}0.105 & \cellcolor{yellow!0}0.118 & \cellcolor{orange!25}0.171 & \cellcolor{green!40}0.707 & \cellcolor{yellow!25}0.359 & \cellcolor{yellow!0}0.108 & \cellcolor{yellow!0}0.083 & \cellcolor{lime!50}0.601 \\
					PP & \cellcolor{green!40}0.744 & \cellcolor{green!40}0.741 & \cellcolor{orange!25}0.202 & \cellcolor{green!40}0.730 & \cellcolor{yellow!25}0.332 & \cellcolor{yellow!25}0.354 & \cellcolor{lime!50}0.531 & \cellcolor{green!40}0.707 \\				
					ADJP & \cellcolor{lime!50}0.543 & \cellcolor{lime!50}0.556 & \cellcolor{orange!25}0.219 & \cellcolor{yellow!25}0.324 & \cellcolor{orange!25}0.263 & \cellcolor{orange!25}0.217 & \cellcolor{yellow!0}0.108 & \cellcolor{lime!50}0.686 \\
					ADVP & \cellcolor{lime!50}0.565 & \cellcolor{lime!50}0.582 & \cellcolor{orange!25}0.187 & \cellcolor{lime!50}0.627 & \cellcolor{yellow!25}0.338 & \cellcolor{yellow!25}0.353 & \cellcolor{orange!25}0.292 & \cellcolor{lime!50}0.655 \\
					\midrule
					Distituent & \cellcolor{yellow!0}0.031 & \cellcolor{yellow!0}0.032 & \cellcolor{yellow!0}0.052 & \cellcolor{yellow!0}0.060 & \cellcolor{yellow!0}0.045 & \cellcolor{yellow!0}0.012 & \cellcolor{yellow!0}0.026 & \cellcolor{yellow!0}0.086 \\
					\midrule
					F1 & 51.1 & 50.9 & 29.5 & 38.6 & 37.4 & 18.6 & 15.0 & 43.1 \\
					\bottomrule
			\end{tabular}}
		\end{center}
		\caption{\label{table:constestanalysis} For each constituency test and each phrase type XP, we report the fraction of XPs in the PTB development set that pass the constituency test, where we judge each span individually and threshold the grammaticality judgment at 0.5. We also report F1 (calculated over all spans). Before refinement, coordination consistently fires for all categories but also for almost half of the distituents. The other tests behave roughly as expected; for example, the NP proforms (``ones'' and ``it'') fire for NPs, while the VP proform (``did so'') fires for VPs. After refinement, coordination no longer fires for distituents, and all of the tests have higher F1. In addition, the proforms now fire for a much wider range of phrase types. See the appendix for a grayscale version.}
	\end{table}
	
	\begin{figure*}[t]
		\centering
		\resizebox{0.99\linewidth}{!}{
		
		\begin{tabular}{lr}
		
		\begin{tabular}{lrr}
		& \multirow{2}{*}{\begin{tikzpicture}
		\tiny
		\tikzset{level distance=10pt}
		\tikzset{sibling distance=-1pt}
		\tikzset{edge from parent/.style=
			{draw,
				edge from parent path={(\tikzparentnode.south)
					-- +(0,0pt)
					-| (\tikzchildnode)}}}
		\Tree [ .1.00 \edge[draw=blue]; [ .0.16 \edge[very thick,draw=red]; [ .0.61 \edge[draw=blue]; Both \edge[draw=blue]; funds ] \edge[very thick,draw=red]; [ .0.34 \edge[very thick,draw=red]; are \edge[very thick,draw=red]; expected ] ] \edge[draw=blue]; [ .0.37 \edge[draw=blue]; [ .0.31 \edge[very thick,draw=red]; to \edge[very thick,draw=red]; [ .0.24 \edge[dashed,draw=blue]; begin \edge[dashed,draw=blue]; operation ] ] \edge[draw=blue]; [ .0.63 \edge[dashed,draw=blue]; [ .0.77 \edge[draw=blue]; [ .0.54 \edge[very thick,draw=red]; around \edge[very thick,draw=red]; March ] \edge[draw=blue]; 1 ] \edge[dashed,draw=blue]; [ .0.18 \edge[dashed,draw=blue]; [ .1.00 \edge[dashed,draw=blue]; , \edge[dashed,draw=blue]; subject ] \edge[dashed,draw=blue]; [ .0.16 \edge[draw=blue]; [ .0.24 \edge[very thick,draw=red]; to \edge[very thick,draw=red]; [ .0.37 \edge[dashed,draw=blue]; [ .0.24 \edge[dashed,draw=blue]; Securities \edge[dashed,draw=blue]; and ] \edge[dashed,draw=blue]; [ .0.36 \edge[dashed,draw=blue]; Exchange \edge[dashed,draw=blue]; Commission ] ] ] \edge[draw=blue]; approval ] ] ] ] ]
		\end{tikzpicture}} \\\\\\\\\\
        (a) Before & \\ 
        Refinement &
		
		\\
		
		\midrule
		
		& \multirow{2}{*}{\begin{tikzpicture}
		\tiny
		\tikzset{level distance=10pt}
		\tikzset{sibling distance=-1pt}
		\tikzset{edge from parent/.style=
			{draw,
				edge from parent path={(\tikzparentnode.south)
					-- +(0,0pt)
					-| (\tikzchildnode)}}}
		\Tree [ .1.00 \edge[draw=blue]; [ .0.27 \edge[very thick,draw=red]; [ .0.57 \edge[draw=blue]; Both \edge[draw=blue]; funds ] \edge[very thick,draw=red]; are ] \edge[draw=blue]; [ .0.29 \edge[draw=blue]; expected \edge[draw=blue]; [ .0.53 \edge[draw=blue]; [ .0.49 \edge[very thick,draw=red]; to \edge[very thick,draw=red]; [ .0.26 \edge[dashed,draw=blue]; begin \edge[dashed,draw=blue]; operation ] ] \edge[draw=blue]; [ .0.84 \edge[dashed,draw=blue]; [ .0.99 \edge[draw=blue]; around \edge[draw=blue]; [ .0.59 \edge[draw=blue]; March \edge[draw=blue]; 1 ] ] \edge[dashed,draw=blue]; [ .0.55 \edge[dashed,draw=blue]; [ .1.00 \edge[dashed,draw=blue]; , \edge[dashed,draw=blue]; subject ] \edge[dashed,draw=blue]; [ .0.24 \edge[draw=blue]; to \edge[draw=blue]; [ .0.57 \edge[draw=blue]; [ .0.37 \edge[dashed,draw=blue]; Securities \edge[dashed,draw=blue]; [ .0.09 \edge[dashed,draw=blue]; and \edge[dashed,draw=blue]; Exchange ] ] \edge[draw=blue]; [ .0.35 \edge[dashed,draw=blue]; Commission \edge[dashed,draw=blue]; approval ] ] ] ] ] ] ] ]
		\end{tikzpicture}} \\
		\\\\\\\\\\
		(b) After \\
		Refinement \\ 
		
		\midrule

		& \multirow{2}{*}{\begin{tikzpicture}
		\tiny
		\tikzset{level distance=8pt}
		\tikzset{sibling distance=-1pt}
		\tikzset{edge from parent/.style=
			{draw,
				edge from parent path={(\tikzparentnode.south)
					-- +(0,0pt)
					-| (\tikzchildnode)}}}
		\Tree [ .$\;$ \edge[draw=blue]; [ \edge[very thick,draw=red]; [ \edge[draw=blue]; Both \edge[draw=blue]; funds ] \edge[very thick,draw=red]; [ \edge[very thick,draw=red]; are \edge[very thick,draw=red]; [ \edge[very thick,draw=red]; expected \edge[very thick,draw=red]; [ \edge[very thick,draw=red]; to \edge[very thick,draw=red]; [ \edge[dashed,draw=blue]; begin \edge[dashed,draw=blue]; [ \edge[dashed,draw=blue]; operation \edge[dashed,draw=blue]; [ \edge[draw=blue]; around \edge[draw=blue]; [ \edge[draw=blue]; March \edge[draw=blue]; 1 ] ] ] ] ] ] ] ] \edge[draw=blue]; [ \edge[dashed,draw=blue]; [ \edge[dashed,draw=blue]; , \edge[dashed,draw=blue]; subject ] \edge[dashed,draw=blue]; [ \edge[draw=blue]; to \edge[draw=blue]; [ \edge[draw=blue]; [ \edge[dashed,draw=blue]; Securities \edge[dashed,draw=blue]; [ \edge[dashed,draw=blue]; and \edge[dashed,draw=blue]; Exchange ] ] \edge[draw=blue]; [ \edge[dashed,draw=blue]; Commission \edge[dashed,draw=blue]; approval ] ] ] ] ]
		\end{tikzpicture}}
		
		\\\\\\

		(c) After \\
		Refinement \\
		+ URNNG \\
		
		\midrule
		
		(d) Gold &
		
		\begin{tikzpicture}
		\tiny
		\tikzset{level distance=10pt}
		\tikzset{sibling distance=-1pt}
		\tikzset{edge from parent/.style=
			{draw,
				edge from parent path={(\tikzparentnode.south)
					-- +(0,0pt)
					-| (\tikzchildnode)}}}
		\Tree [ .S [ .NP Both funds ] [ .VP are [ .VP expected [ .VP to [ .VP begin operation [ .PP around [ .NP March 1 ] ] , [ .ADJP subject [ .PP to [ .NP Securities and Exchange Commission approval ] ] ] ] ] ] ] ]

		\end{tikzpicture}\\
		
		\midrule
		
		\end{tabular}
		
		& \begin{tikzpicture}[
          bluedashnode/.style={shape=circle, draw=blue, dashed},
          bluenode/.style={shape=circle, draw=blue},
          rednode/.style={shape=circle, draw=red, line width=2}
        ]
        \matrix {
              \node [bluenode,label=right:Correct bracket] {}; \\
              \node [bluedashnode,label=right:Consistent bracket] {}; \\
              \node [rednode,label=right:Crossing bracket] {}; \\
            };
        \end{tikzpicture}
        
        \end{tabular}
        }
		\caption{\label{fig:exampletrees} Example trees (a) before refinement, (b) after refinement, (c) after refinement + URNNG, and (d) gold, where we use the first PTB train sentence whose F1 was within 1 of the average. Each non-trivial span is labeled with its score under the model, i.e.\ the average grammaticality of its constituency tests. Each span is labeled blue if it is present in the gold, dashed blue if it is consistent (ignoring punctuation), and thick red if it is crossing. After refinement (tree b), the parser makes two mistakes: attaching ``are'' to the subject, and attaching the phrase ``around March ... Commission approval'' one level too high. After refinement + URNNG (tree c), the only mistake is attaching the phrase ``subject to ... Commission approval'' at the top level, which produces four crossing brackets. }
	\end{figure*}
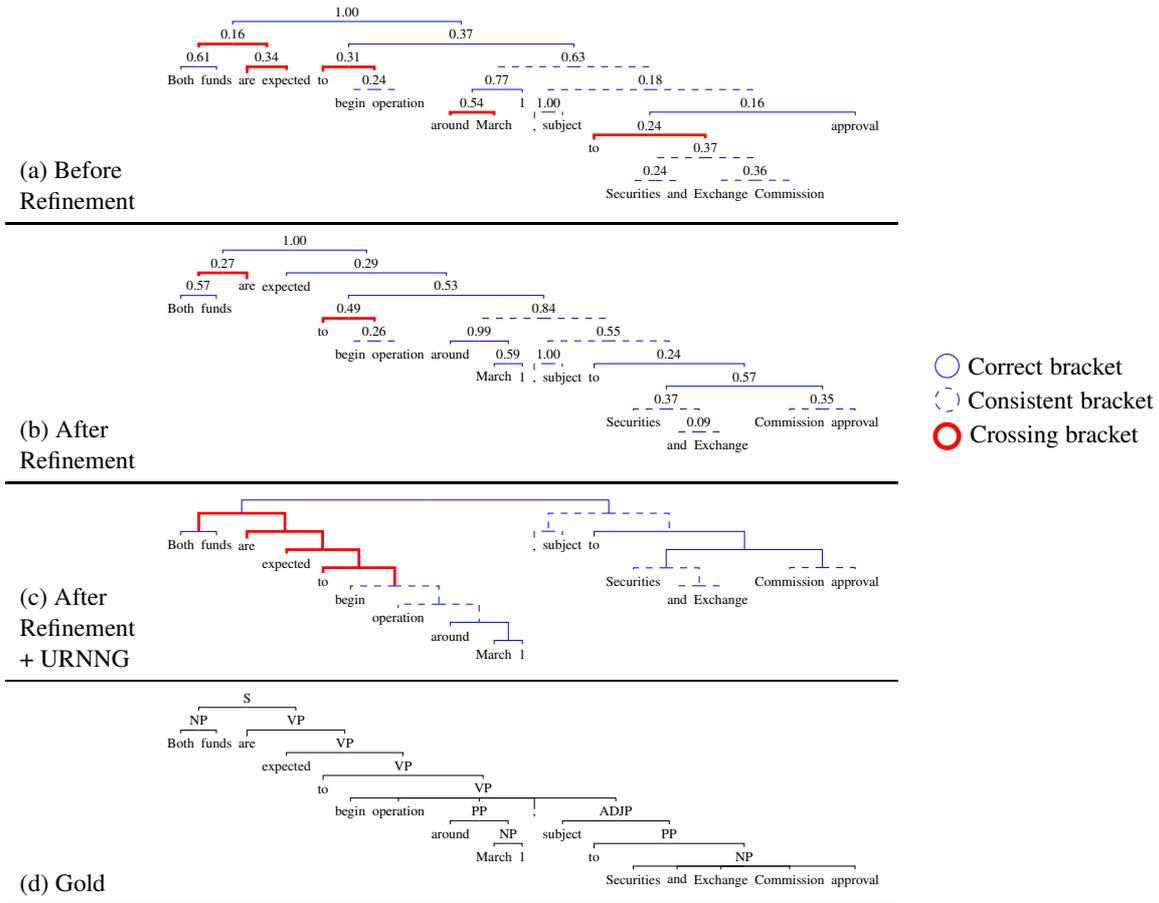
	
	\subsection{Recall by Label}
	First, we compute recall by label for the parser before refinement, after refinement, and after refinement + URNNG, displayed in Table~\ref{table:recallbylabel}. Before refinement, the parser is strongest in ADJPs and ADVPs and weakest for VPs and SBARs. Refinement causes all categories except ADJP and ADVP to receive a boost of about 0.3 in recall. Afterward, URNNG produces a boost for SBAR and VP, resulting in the four categories being above 0.8, except with ADJP and ADVP still both around 0.55. In Section~\ref{section:mistakes}, we analyze the sources of these mistakes in more detail and find that the model is less effective in identifying ADJPs that serve as NP adjuncts (e.g.\ \textit{``}[ \textit{most recent} ] \textit{news}\textit{''}).
	
	\subsection{Analyzing the Constituency Tests}\label{section:analyzingtests}
	To better understand how well each category is covered by constituency tests, in Table~\ref{table:constestanalysis} we display recall per phrase type for each test, along with F1 computed over all spans. Using each test, we judge each span in the PTB development set individually by thresholding the grammaticality judgment at 0.5, and for each phrase type we report the fraction that pass the test. Before refinement, the tests behave roughly as expected. Coordination fires for all phrase types but also half the distituents, while the NP and VP proforms fire for NPs and VPs respectively. Clefting and movement are more mixed, with clefting sometimes firing for all phrase types except VP, and movement sometimes firing for SBARs, PPs, and ADVPs. Interestingly, the individual F1 numbers are all quite low at around 10-20 F1, even though the parser achieves 48.1 F1, suggesting that the constraint of outputting a well-formed tree provides substantial information. After refinement, all of the tests have better F1, potentially because refinement allows the grammar model to use the well-formedness constraint to improve its span judgments (see Section~\ref{section:em}). In particular, we find that coordination no longer has false positives, and clefting exhibits greatly improved recall. We also see that the proform substitution tests now fire for a wider range of phrase types; for example, ``did so'' now fires for 70\% of SBARs, VPs, PPs, and ADVPs, even though it was originally a VP substitution.
	
	\begin{table}[t] 
		\begin{center}
			\resizebox{\linewidth}{!}{\begin{tabular}{llcc}
					\toprule
					\multicolumn{4}{l}{Common mistakes after refinement} \\
				    &  & Percentage & $\Delta$ in \# mistakes \\
					Parts of speech  & Example & of mistakes & after URNNG \\
					\midrule
					PRP VBD/P/Z & [ \textit{they 're} ] \textit{squaring off} & 1.72\% & -81.0\% \\
					IN NN(S) & [ \textit{in letters} ] \textit{to the agency} & 1.07\% & -57.6\% \\
					CD NN(S) & \textit{about} [ \textit{1,200 cars} ] & 1.06\% & \phantom{0}+4.4\% \\
					IN DT NN(S) & [ \textit{in an effort} ] \textit{to streamline} & 0.99\% & -74.7\% \\
					TO VB & [ \textit{to work} ] \textit{a lot} & 0.93\% & -95.0\% \\
					\midrule
					\midrule
					\multicolumn{4}{l}{Common mistakes after refinement + URNNG} \\
				    &  & Percentage & $\Delta$ in \# mistakes \\
					Parts of speech  & Example & of mistakes & after URNNG \\
					\midrule
					CD NN(S) & \textit{about} [ \textit{1,200 cars} ] & 1.51\% & \phantom{0}+4.4\% \\
					JJ NN(S) & \textit{socially} [ \textit{responsible companies} ] & 0.69\% & +47.0\% \\
					IN NN(S) & [ \textit{in letters} ] \textit{to the agency} & 0.61\% & -57.6\% \\
					NN(S) IN NN(S) & [ \textit{plenty of reasons} ] \textit{to stay} & 0.57\% & -27.3\% \\
					NNP VBD/P/Z & \textit{Mr.} [ \textit{Lane said} ] & 0.47\% & -21.2\% \\
					\bottomrule
			\end{tabular}}
		\end{center}
		\caption{\label{table:mistakes} The five most common crossing brackets categorized by part-of-speech, computed on the first 5,000 sentences in the PTB training set. We also report percentage of crossing predicted brackets (i.e.\ mistakes) that fall under that category, as well as the change in the number of mistakes after adding URNNG. We group (VBD, VBP, VBZ) (past, present, present 3rd-person) and (NN, NNS) (noun, noun plural). We find that the model commonly makes the following mistakes: (1) bracketing the verb with the subject, (2) in a nested PP, attaching the inner PP outside, (3) grouping the cardinal or adjective with the noun instead of with its adverb, and (4) bracketing ``to + infinitive.'' After URNNG, each of the mistakes are corrected except (3).}
	\end{table}
	
	\subsection{Common Mistakes}\label{section:mistakes}
    In Table~\ref{table:mistakes}, we show the most common crossing brackets predicted by the parser, where for analysis we categorize the brackets by part-of-speech. We find that the model after refinement commonly makes the following mistakes, and we suggest possible explanations for each:
    \begin{enumerate}
        \item Bracketing the verb with the subject:
        
        [ \textit{they 're} ] \textit{squaring off}
    
        As shown in Table~\ref{table:constestanalysis}, there is less support for VPs via consituency tests. This observation is also reflected in the example trees in Figure~\ref{fig:exampletrees}, where the VPs have consistently lower scores. Therefore, while the parser usually chooses to bracket VPs (achieving 0.682 recall, as shown in Table~\ref{table:recallbylabel}), there seem to be cases in which it prefers the [ subject verb ] bracketing.
    
        \item In a nested PP, attaching the inner PP outside the outer PP:
    
        [ \textit{in letters} ] \textit{to the agency}
    
        The spans resulting from incorrect attachments still tend to produce grammatical constituency tests (e.g.\ \textit{``they argue }[\textit{ in letters }]\textit{ to the agency that ...''} $\to$ \textit{``in letters , they argue to the agency that ...''}).
    
        \item Grouping cardinals and adjectives with the noun, instead of with the adverb:
    
        \textit{about} [ \textit{1,200 cars} ]
        
        This span passes some constituency tests, like \textit{``about \{it, ones\},''} while none of the tests except coordination accept \textit{``about 1,200.''}
        
        \item Bracketing ``to + infinitive'':
    
        \textit{they want} [ \textit{to work} ] \textit{a lot}
        
        Infinitive VPs (e.g.\ \textit{``work a lot''}) typically don't pass any of our tests except coordination, while ``to + infinitive'' is often replaceable by a noun proform, like \textit{``they want it a lot.''}
    \end{enumerate}
    After URNNG, the VP errors (1 and 4) are corrected almost completely, while the PP attachment error also decreases in frequency by about half. In contrast, the ADJP error (3) is exacerbated, with [ CD NN ] and [ JJ NN ] incorrect bracketings increasing by 4.4\% and 47.0\% (Table~\ref{table:mistakes}). Therefore, URNNG is effective in correcting many but not all of the parser's systematic errors, suggesting paths for future improvement, e.g.\ by adding tests that fire for currently missing brackets.
    
    \subsection{Example Trees}
	
	Finally, to qualitatively understand the parser's performance, in Figure~\ref{fig:exampletrees} we display the trees before refinement, after refinement, and after refinement + URNNG for the sentence \textit{``Both funds are expected to begin operation around March 1 , subject to Securities and Exchange Commission approval.''} To produce a representative example, we selected this sentence by choosing the first sentence in PTB train whose F1 was within 1 of the average. Comparing the trees before and after refinement, the parser corrects two mistakes, \textit{``}[\textit{ around March }]\textit{ 1''} and \textit{``}[\textit{ to Securities and Exchange Commission }]\textit{ approval,''} which both involve bracketing the preposition with part of its NP complement. As a result, ignoring punctuation and binarization, the parser after refinement makes only two mistakes: attaching \textit{``are''} to the subject, and attaching the phrases \textit{``around March''} and \textit{``subject to ... Commission approval''} one level too high. After URNNG, the first mistake is corrected, such that the only mistake is in the attachment of \textit{``subject to ... Commission approval''} (but because it attaches this phrase very high, this mistake produces four crossing brackets). This example provides some characterization of each step's improvement to the predicted trees.
	
	\section{Conclusion}
	In this paper, we showed that using constituency tests to parse sentences is an effective approach, achieving strong performance for unsupervised parsing. Furthermore, we used the interpretability of constituency tests to highlight and explain the parser's strengths and shortcomings, like the ``[ subject verb ]'' and ``adverb [ adjective noun ]'' misbracketings, revealing potential next steps for improvement. Therefore, we see parsing via constituency tests as a promising new approach with both strong results and many open questions.
	
	\section*{Acknowledgments}

    This research was supported by the National Science Foundation under Grant No.\ 1618460 and by DARPA under the LwLL program / Grant No.\ FA8750-19-1-0504. This work used the Savio computational cluster provided by the Berkeley Research Computing program at the University of California, Berkeley.
	
	\bibliographystyle{acl_natbib}
	\bibliography{emnlp2020}
	
	\newpage
	\appendix
	\section{Appendix}
	\subsection{Optimization Hyperparameters and Other Training Details}
	For both real/fake training and refinement, we use a learning rate of $3 \times 10^{-5}$ with Adam~\citep{kingma-ba-2015-adam} hyperparameters $\beta = (0.9, 0.999)$, $\epsilon = 10^{-6}$ and linear learning rate warmup for the first 10\% of the training data. For real/fake training, each batch contains 32 real and 32 fake sentences, while for refinement we parse a batch of 32 sentences for each gradient step. We did not perform any hyperparameter search.
	
	We fine-tuned the RoBERTa base model, which has 125M parameters, and we performed classification for sentences by applying a linear layer and softmax to the $\texttt{[CLS]}$ embedding. 
	
	For real/fake training, we used a single Nvidia K80 with 12GB RAM, which took about 3 days to run for 5 million sentences. For refinement, we either used a single Quadro 8000 with 48GB RAM, which took about 1 day to run, or a single Nvidia K80, which took about 6 days to run.
	
    For URNNG, we used the default hyperparameters in the \citet{kim-etal-2019-unsupervised} github. Specifically, we used a batch size of 16, and we performed 18 epochs of supervised RNNG training with a learning rate of 0.0001, and 10 epochs of unsupervised fine-tuning with a learning rate of 0.1. Other optimization details can be found in the original paper~\citep{kim-etal-2019-unsupervised}. We used a single Quadro 6000 with 24GB RAM, which took about 3 days.
    
    As our data, we used the first 5M sentences from the English Gigaword corpus~\citep{graff-cieri-2003-gigaword} for real/fake training, and we used the standard train/development/test splits (sections 02-21, 22, 23) of the Penn Treebank for parsing~\citep{marcus-etal-1993-building}, which have 39832, 1700, and 2416 examples, respectively. Both datasets are already tokenized. For preprocessing, we converted all letters to lowercase and removed quotation marks and any ending punctuation. 
    
    \subsection{Some Ablations of the Refinement Procedure}
    Having analyzed the output of our parser, next we describe some ablations to determine how much of the performance is due to constituency tests versus the refinement procedure. 
    
    First, if we ablate the refinement procedure (Table~\ref{table:ptb}), the initial parser still performs quite well -- it is much better than right-branching and relatively close in performance to current methods. We can also try ablating the constituency tests. Specifically, following the suggestion of an anonymous reviewer, we randomly initialized a Roberta-based span classification parser and performed refinement of the span scores (Section~\ref{section:em}). The resulting parser did not achieve very high accuracy (initial F1: 11.95, final F1: 12.33; F1 is computed including punctuation). These ablations suggest that constituency tests are the main driving force behind our method. We discuss a few possible reasons below.\phantom{\cite{teng-zhang-2018-two}}

    First, because the refinement method has the effect of enforcing self-consistency, the initialization is important, and constituency tests are important for the initialization.
    
    Next, the refinement procedure itself also relies heavily on constituency tests because the gradient step involves maximizing the grammaticality of constituency tests for spans within the imputed trees. In particular, all span judgments originate from grammaticality judgments, and the only parameters are those in the grammaticality model. Therefore, the procedure exploits the fact that grammaticality and constituency are linked.

    \begin{table}[t] 
		\begin{center}
			\resizebox{\linewidth}{!}{\begin{tabular}{lccccccccc}
					\toprule
					& \multicolumn{2}{c|}{Clefting} & \multicolumn{3}{c|}{Proform Substitution} & \multicolumn{2}{c|}{Movement} & Coord-\\
					& is & \multicolumn{1}{c|}{was} & ones & did so & \multicolumn{1}{c|}{it} & front & \multicolumn{1}{c|}{end} & ination \\
					\midrule
					\midrule
					\multicolumn{2}{l}{Before Refinement} \\
					\midrule
					SBAR & \cellcolor{gray!15}0.294 & \cellcolor{gray!15}0.260 & \cellcolor{gray!0}0.113 & \cellcolor{gray!0}0.130 & \cellcolor{gray!0}0.146 & \cellcolor{gray!15}0.279 & \cellcolor{gray!30}0.319 & \cellcolor{gray!75}0.942 \\
					NP & \cellcolor{gray!30}0.353 & \cellcolor{gray!30}0.347 & \cellcolor{gray!45}0.458 & \cellcolor{gray!15}0.178 & \cellcolor{gray!45}0.555 & \cellcolor{gray!0}0.055 & \cellcolor{gray!0}0.048 & \cellcolor{gray!75}0.934 \\
					VP & \cellcolor{gray!0}0.091 & \cellcolor{gray!0}0.089 & \cellcolor{gray!0}0.067 & \cellcolor{gray!45}0.479 & \cellcolor{gray!0}0.127 & \cellcolor{gray!0}0.060 & \cellcolor{gray!0}0.144 & \cellcolor{gray!75}0.944 \\
					PP & \cellcolor{gray!30}0.427 & \cellcolor{gray!30}0.412 & \cellcolor{gray!15}0.238 & \cellcolor{gray!15}0.165 & \cellcolor{gray!15}0.154 & \cellcolor{gray!30}0.308 & \cellcolor{gray!45}0.606 & \cellcolor{gray!75}0.906 \\
					ADJP & \cellcolor{gray!30}0.383 & \cellcolor{gray!30}0.361 & \cellcolor{gray!15}0.286 & \cellcolor{gray!15}0.241 & \cellcolor{gray!30}0.346 & \cellcolor{gray!0}0.127 & \cellcolor{gray!15}0.172 & \cellcolor{gray!75}0.911 \\
					ADVP & \cellcolor{gray!30}0.396 & \cellcolor{gray!30}0.395 & \cellcolor{gray!0}0.143 & \cellcolor{gray!15}0.185 & \cellcolor{gray!15}0.198 & \cellcolor{gray!15}0.290 & \cellcolor{gray!30}0.307 & \cellcolor{gray!75}0.893 \\
					\midrule
					Distituent & \cellcolor{gray!0}0.066 & \cellcolor{gray!0}0.063 & \cellcolor{gray!15}0.184 & \cellcolor{gray!0}0.098 & \cellcolor{gray!0}0.123 & \cellcolor{gray!0}0.033 & \cellcolor{gray!0}0.052 & \cellcolor{gray!45}0.456 \\
					\midrule
					F1 & 20.8 & 20.9 & 10.9 & 13.2 & 17.8 & 12.3 & 13.6 & 16.1 \\
					\midrule
					\midrule
					\multicolumn{2}{l}{After Refinement} \\
					\midrule
					SBAR & \cellcolor{gray!15}0.237 & \cellcolor{gray!15}0.223 & \cellcolor{gray!15}0.237 & \cellcolor{gray!75}0.770 & \cellcolor{gray!30}0.374 & \cellcolor{gray!15}0.250 & \cellcolor{gray!15}0.225 & \cellcolor{gray!45}0.539 \\
					NP & \cellcolor{gray!75}0.718 & \cellcolor{gray!75}0.712 & \cellcolor{gray!45}0.571 & \cellcolor{gray!30}0.428 & \cellcolor{gray!45}0.539 & \cellcolor{gray!0}0.063 & \cellcolor{gray!0}0.035 & \cellcolor{gray!75}0.792 \\
					VP & \cellcolor{gray!0}0.105 & \cellcolor{gray!0}0.118 & \cellcolor{gray!15}0.171 & \cellcolor{gray!75}0.707 & \cellcolor{gray!30}0.359 & \cellcolor{gray!0}0.108 & \cellcolor{gray!0}0.083 & \cellcolor{gray!45}0.601 \\
					PP & \cellcolor{gray!75}0.744 & \cellcolor{gray!75}0.741 & \cellcolor{gray!15}0.202 & \cellcolor{gray!75}0.730 & \cellcolor{gray!30}0.332 & \cellcolor{gray!30}0.354 & \cellcolor{gray!45}0.531 & \cellcolor{gray!75}0.707 \\				
					ADJP & \cellcolor{gray!45}0.543 & \cellcolor{gray!45}0.556 & \cellcolor{gray!15}0.219 & \cellcolor{gray!30}0.324 & \cellcolor{gray!15}0.263 & \cellcolor{gray!15}0.217 & \cellcolor{gray!0}0.108 & \cellcolor{gray!45}0.686 \\
					ADVP & \cellcolor{gray!45}0.565 & \cellcolor{gray!45}0.582 & \cellcolor{gray!15}0.187 & \cellcolor{gray!45}0.627 & \cellcolor{gray!30}0.338 & \cellcolor{gray!30}0.353 & \cellcolor{gray!15}0.292 & \cellcolor{gray!45}0.655 \\
					\midrule
					Distituent & \cellcolor{gray!0}0.031 & \cellcolor{gray!0}0.032 & \cellcolor{gray!0}0.052 & \cellcolor{gray!0}0.060 & \cellcolor{gray!0}0.045 & \cellcolor{gray!0}0.012 & \cellcolor{gray!0}0.026 & \cellcolor{gray!0}0.086 \\
					\midrule
					F1 & 51.1 & 50.9 & 29.5 & 38.6 & 37.4 & 18.6 & 15.0 & 43.1 \\
					\bottomrule
			\end{tabular}}
		\end{center}
		\caption{A grayscale version of Table~\ref{table:constestanalysis}, where higher numbers are shaded with darker shades of gray.}
	\end{table}
	
\end{document}